\documentclass[letterpaper]{article} 
\usepackage{aaai2027}  
\usepackage[hyphens]{url}  
\usepackage{graphicx} 
\usepackage{natbib}  
\usepackage{caption} 
\usepackage{algorithm}
\usepackage{algorithmic}

\usepackage{newfloat}
\usepackage{listings}
\DeclareCaptionStyle{ruled}{labelfont=normalfont,labelsep=colon,strut=off} 
\floatstyle{ruled}
\newfloat{listing}{tb}{lst}{}
\floatname{listing}{Listing}

\usepackage{booktabs}
\usepackage{array}
\usepackage{amsmath}
\usepackage{amssymb}

\newcolumntype{L}[1]{>{\raggedright\arraybackslash}p{#1}}

\title{From Passive Video to Editable Experience: Physically Grounded Experience Synthesis for Embodied Intelligence}
\author{
    Jia Luo
}
\affiliations{
    Huazhong University of Science and Technology\\
    u202317016@hust.edu.cn
}

\begin{document}

\maketitle

\begin{abstract}
The key bottleneck in embodied AI is not model architecture but data. Although billions of human manipulation videos exist online, robots cannot directly learn from them due to the embodiment gap between human morphology and robot hardware. We introduce Pegasus, a low-resource framework that bridges this gap by translating human demonstrations into robot-learnable data through structured knowledge transfer. Instead of relying on raw video prompts, Pegasus constructs a graph-based intermediate representation: a Task Graph extracted from human videos is transformed through Affordance and Constraint Graphs into a Robot Planning Graph for robot-conditioned video generation. A hierarchical affordance latent space models the relationship between object states, affordances, and tasks, enabling generalization beyond object identities. A closed-loop physics verifier further filters invalid generations using kinematic feasibility, collision constraints, and joint limits. We evaluate Pegasus across a range of egocentric manipulation benchmarks, including GTEA Gaze+ and EPIC-KITCHENS-100, and diverse robot embodiments, assessing Task Correctness, Executability, State Consistency, and Learnability. Results demonstrate reliable cross-embodiment translation and show that robot data generation can be reframed from a hardware collection problem into a scalable, low-resource knowledge transfer problem.

\end{abstract}


\section{Introduction}

Learning general-purpose robot manipulation requires diverse, large-scale
training data. Current data acquisition paradigms each have key limitations.
Real-robot collection~\cite{brohan2023rt1roboticstransformerrealworld,vuong2023open}
produces high-quality demonstrations but is expensive and difficult to scale.
Physics simulation~\cite{todorov2012mujoco,makoviychuk2021isaacgymhighperformance}
provides unlimited data but suffers from the sim-to-real gap
for vision-based policies~\cite{aljalbout2025realitygaproboticschallenges}.
Learning from human videos~\cite{nair2022r3muniversalvisualrepresentation,ma2023vipuniversalvisualreward}
leverages abundant internet-scale data, yet existing approaches primarily learn
visual representations rather than executable robot demonstrations.

Meanwhile, large human-video datasets such as EPIC-Kitchens~\cite{damen2022rescaling},
Ego4D~\cite{grauman2024ego4d}, DexYCB~\cite{chao2021dexycb}, and
H2O~\cite{kwon2021h2o} capture task goals, object interactions, and state
transitions. The fundamental challenge is the \emph{embodiment gap}: humans and
robots differ in morphology, kinematics, and action spaces, so human
demonstrations do not transfer directly to robot execution.

We argue that bridging this gap requires transferring \emph{structured
experience} rather than pixels. Instead of matching human and robot appearances,
a system should extract task semantics, model object affordances, enforce robot
constraints, and generate demonstrations that are executable for downstream
policy learning.

To this end, we propose \textsc{Pegasus}, a framework for cross-embodiment
translation from human videos to robot demonstrations. Pegasus represents human
manipulation with structured graphs, learns a hierarchical Affordance Latent
that captures object interaction semantics instead of object identity, and
integrates closed-loop physics verification to ensure kinematic feasibility.
These components enable scalable synthesis of robot-learnable demonstrations
from internet videos while preserving task semantics across embodiments.

Our experiments demonstrate that Pegasus consistently improves task correctness,
physical executability, affordance generalization to unseen objects, and
downstream robot policy learning over strong baselines.
Together, these results validate structured experience transfer as a practical
alternative to collecting every demonstration on physical hardware.

Our contributions are summarized as follows:

\begin{itemize}
    \item \textbf{Transfer formulation.} We recast internet human videos as a scalable source of robot training data and formulate acquisition as cross-embodiment knowledge transfer.
    \item We propose a \textbf{graph-structured cross-embodiment translation} framework that explicitly models task semantics, affordances, embodiment constraints, and robot planning.
    \item We introduce a \textbf{hierarchical Affordance Latent} that improves generalization by organizing representations according to object states, affordances, and tasks.
    \item We develop a \textbf{closed-loop physics verification} mechanism that enforces kinematic feasibility through iterative verification and regeneration.
    \item Extensive experiments demonstrate improved task correctness, unseen-object generalization, and downstream policy learning compared with existing approaches.
\end{itemize}
\section{Related Work}

\subsection{Embodied Data Acquisition}

Robot learning has been defined by its data paradigms. \textit{Simulation}~\cite{todorov2012mujoco,makoviychuk2021isaacgymhighperformance} drives RL progress but the sim-to-real gap persists.
\textit{Real-robot collection}~\cite{brohan2023rt1roboticstransformerrealworld,vuong2023open,collaboration2023open} yields impressive results
but at prohibitive hardware cost. \textit{Learning from human video}~\cite{nair2022r3muniversalvisualrepresentation,ma2023vipuniversalvisualreward,bahl2023affordances}
extracts reusable representations but cannot generate executable robot trajectories.
\textsc{Pegasus} introduces a fourth paradigm: \textit{translating} human experience into
robot-learnable data through structured cross-embodiment knowledge transfer.

\subsection{Video Generation for Robotics}

Video diffusion models~\cite{ho2022video,blattmann2023stablevideodiffusionscaling} and video inpainting
methods~\cite{zhou2023propainter,ouyang2024codefcontentdeformationfields} achieve photorealism.
Recent works apply generation to embodied data: \textsc{EgoScale} uses video diffusion
for augmentation (GPU-dependent); \textsc{GenAug} operates within simulation.
\textsc{UniPi}~\cite{du2023learning} and \textsc{Susie}~\cite{black2024zero} explore
video as a policy interface but do not address cross-embodiment transfer.
\textsc{Pegasus} differs fundamentally: generation is conditioned on a \textit{graph
representation} of task structure, not raw pixels or text prompts;
our quality metric is \textit{policy improvement}, not visual fidelity.

\subsection{Cross-Embodiment Transfer}

Cross-embodiment learning~\cite{wang2024crossembodimentrobotmanipulationskill,zhu2023learninggeneralizablemanipulationpolicies} studies policy transfer
across robot morphologies. \textsc{Octo}~\cite{mees2024octo} and \textsc{OpenVLA}~\cite{kim2024openvlaopensourcevisionlanguageactionmodel}
build generalist policies across embodiments but require robot data for training.
\textsc{VRB}~\cite{bahl2023affordances} extracts affordances from human video
but does not translate them to executable robot trajectories.
\textsc{Pegasus} uniquely bridges \textit{human video} to \textit{robot-executable data}
through graph-structured task translation and affordance-centric representation.

\subsection{Graph Representations for Manipulation}

Graph-based representations have become a fundamental abstraction for robot manipulation, providing structured priors for task planning, scene understanding, and action reasoning. Task graphs~\cite{konidaris2018skills} encode high-level procedural structures for long-horizon planning, while scene graphs~\cite{krishna2017visual} and affordance graphs~\cite{bahl2023affordances} capture object relationships and actionable interactions within an environment. Recent work further extends these representations to 3D scene graphs, enabling spatial-semantic reasoning over complex indoor environments~\cite{roberts2021hypersim}. Building on these advances, \textsc{Pegasus} leverages graph representations as an intermediate abstraction for cross-embodiment translation, bridging human demonstrations and robot-executable trajectories. Unlike prior work, which primarily employs graphs for perception, planning, or semantic reasoning, \textsc{Pegasus} uses graph-structured representations to facilitate embodiment transfer within a video generation pipeline.

\section{Method: \textsc{Pegasus}}

\subsection{Problem Formulation}

Given a human egocentric demonstration
$V_h = \{I_1, \ldots, I_T\}$ and a target robot embodiment $R$ parameterized by
its kinematic structure, joint limits, end-effector specification, and workspace
constraints, our objective is to synthesize a robot demonstration
\begin{equation}
    D_r = f(V_h, R),
\end{equation}
that satisfies three requirements. First, the generated demonstration should
preserve the underlying task semantics, including sub-goal ordering, object
interactions, and causal dependencies observed in the human demonstration.
Second, all generated motions must satisfy embodiment-specific kinematic and
geometric constraints. Finally, the synthesized demonstrations should serve as
effective supervision for downstream visuomotor policy learning.

Unlike prior work that directly conditions a video generation model using
textual prompts, \textsc{Pegasus} formulates cross-embodiment translation as
structured graph reasoning followed by physics-constrained rendering. This
decomposition explicitly separates embodiment-invariant task representations
from embodiment-dependent execution.

\subsection{Overview}

\textsc{Pegasus} consists of five stages,
\begin{equation}
    V_h \rightarrow \mathcal{G}_T \rightarrow \mathcal{Z}
    \rightarrow \mathcal{G}_P \rightarrow D_r,
\end{equation}
where $\mathcal{G}_T$ denotes a task graph extracted from human demonstrations,
$\mathcal{Z}$ is a hierarchical affordance latent, and $\mathcal{G}_P$ is an
executable robot planning graph.

Rather than mapping videos directly to robot trajectories, \textsc{Pegasus}
progressively converts human demonstrations into embodiment-aware
representations. Task structure and object affordances are assumed to remain
invariant across embodiments, whereas motion trajectories depend on robot
morphology. Embodiment transfer therefore becomes a structured graph
transformation problem rather than an appearance generation problem.

\begin{figure*}[t]
\centering
\includegraphics[width=0.95\textwidth]{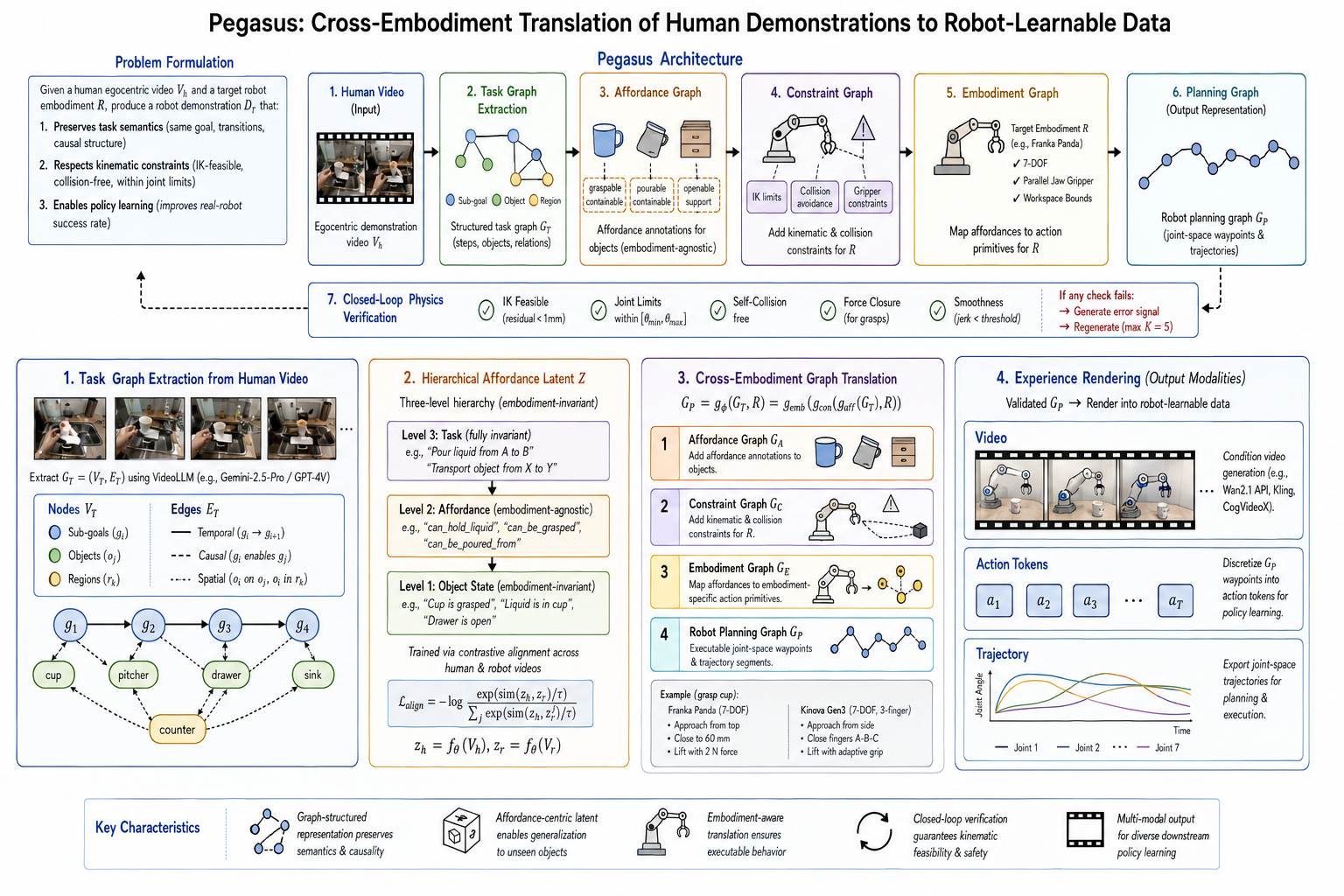}
\caption{\textbf{\textsc{Pegasus} Architecture.} Human demonstrations are transformed into task graphs, encoded in a hierarchical affordance latent space, translated into embodiment-specific robot planning graphs, verified against physical constraints, and rendered as robot-learning experiences.}
\label{fig:pipeline}
\end{figure*}

Figure~\ref{fig:pipeline} illustrates the complete \textsc{Pegasus} framework.

\subsection{Task Graph Extraction}
\label{sec:graph}

Given a human demonstration, \textsc{Pegasus} first constructs a task graph
\begin{equation}
    \mathcal{G}_T = (\mathcal{V}, \mathcal{E}),
\end{equation}
whose nodes represent semantic entities,
\begin{equation}
    \mathcal{V} = \{g_i, o_j, r_k\},
\end{equation}
including sub-goals, manipulated objects, and spatial regions. Edges encode
three complementary relations,
\begin{equation}
    \mathcal{E} = \{e_{\mathrm{temp}}, e_{\mathrm{cause}},
    e_{\mathrm{spatial}}\},
\end{equation}
corresponding to temporal ordering, causal dependency, and spatial interaction.

Rather than reasoning directly over image pixels, \textsc{Pegasus} performs all
subsequent reasoning over this structured graph, which explicitly preserves
task topology while remaining independent of robot morphology. The graph is
extracted using a structured vision-language parser that jointly predicts
semantic actions, manipulated objects, spatial relationships, and temporal
dependencies from video observations.

\subsection{Hierarchical Affordance Latent}
\label{sec:latent}

A central question is determining which representation should remain invariant
across embodiments. Robot poses are embodiment-dependent. Human skeletons are
likewise morphology-specific. Object identities alone fail to generalize to
unseen instances. \textsc{Pegasus} instead introduces a hierarchical affordance
latent
\begin{equation}
    \mathcal{Z} = \{\mathcal{Z}_s, \mathcal{Z}_a, \mathcal{Z}_t\},
\end{equation}
organized from object states to affordances and finally task semantics.

The object-state level captures the physical state of the environment,
\begin{equation}
    \mathcal{Z}_s = \Phi_s(\mathcal{G}_T),
\end{equation}
encoding properties such as whether an object is grasped, opened, or displaced.
The affordance level abstracts object-specific observations into manipulation
capabilities,
\begin{equation}
    \mathcal{Z}_a = \Phi_a(\mathcal{Z}_s),
\end{equation}
including graspability, containment, support, or pourability. Finally, the task
level represents embodiment-independent intentions,
\begin{equation}
    \mathcal{Z}_t = \Phi_t(\mathcal{Z}_a),
\end{equation}
such as transporting, stacking, or pouring. This hierarchy progressively
removes embodiment-specific information while preserving causal task semantics.

\paragraph{Cross-Embodiment Alignment.}
To align human and robot demonstrations within the shared affordance space,
\textsc{Pegasus} optimizes a contrastive objective,
\begin{equation}
    \mathcal{L}_{\mathrm{align}}
    = -\log
    \frac{\exp(\operatorname{sim}(z_h,z_r)/\tau)}
    {\sum_j \exp(\operatorname{sim}(z_h,z_j)/\tau)},
\end{equation}
where positive pairs correspond to demonstrations accomplishing identical tasks
under different embodiments. Unlike appearance-based alignment, the objective
encourages geometric consistency in affordance space, allowing semantically
equivalent manipulations to converge despite visual differences.

\subsection{Cross-Embodiment Graph Translation}
\label{sec:graph_trans}

Once the shared latent representation has been established, \textsc{Pegasus}
progressively transforms the task graph into an executable robot plan. Rather
than directly predicting robot trajectories, the transformation is decomposed
into three graph operators,
\begin{equation}
    \mathcal{G}_T \rightarrow \mathcal{G}_A
    \rightarrow \mathcal{G}_C \rightarrow \mathcal{G}_P.
\end{equation}

The affordance graph augments semantic nodes with manipulation affordances. The
constraint graph incorporates embodiment-dependent feasibility, including grasp
types, inverse kinematics, collision constraints, and workspace limits. Finally,
the planning graph converts feasible manipulation primitives into executable
robot trajectories represented as motion waypoints and temporal transitions.

The complete translation is parameterized as
\begin{equation}
    \mathcal{G}_P = g_{\phi}(\mathcal{G}_T, R),
\end{equation}
where
\begin{equation}
    g_{\phi} = g_{\mathrm{emb}} \circ g_{\mathrm{con}}
    \circ g_{\mathrm{aff}}.
\end{equation}
This modular formulation explicitly disentangles semantic reasoning from
embodiment adaptation, allowing the same task graph to be instantiated across
heterogeneous robotic platforms.

\subsection{Physics-Constrained Verification}
\label{sec:physics}

Visual plausibility alone does not guarantee executable robot behavior.
\textsc{Pegasus} therefore incorporates a closed-loop verification module that
validates generated trajectories with respect to inverse kinematics, joint
limits, collision avoidance, grasp stability, and trajectory smoothness. Let
$\Psi(\mathcal{G}_P)$ denote the verification operator. If any constraint is
violated,
\begin{equation}
    \Psi(\mathcal{G}_P) = 0,
\end{equation}
the verifier produces structured error signals
\begin{equation}
    \epsilon = \operatorname{Verify}(\mathcal{G}_P),
\end{equation}
which are fed back into the planning module,
\begin{equation}
    \mathcal{G}_P^{(k+1)} =
    g_{\phi}(\mathcal{G}_T, R, \epsilon^{(k)}),
\end{equation}
forming an iterative refinement process until all constraints are satisfied.
This closed-loop optimization converts embodiment adaptation into a constraint
satisfaction problem rather than one-shot trajectory prediction.

\subsection{Experience Rendering}
\label{sec:render}

The verified planning graph is finally rendered into task-specific supervision
suitable for downstream robot learning. Depending on the target policy
architecture, \textsc{Pegasus} supports three output modalities,
\begin{equation}
    D_r \in \{D_{\mathrm{video}}, D_{\mathrm{action}},
    D_{\mathrm{trajectory}}\}.
\end{equation}
Video demonstrations supervise vision-language policies, discrete action tokens
train autoregressive action models, and continuous trajectories provide
executable robot motions for classical control pipelines. Because rendering is
conditioned on executable planning graphs rather than textual descriptions,
generated demonstrations preserve both semantic consistency and physical
feasibility across heterogeneous robot embodiments.


\section{Experiments}
\label{sec:experiments}

Our experiments are organized around four questions that an ICLR/AAAI reviewer
would ask about any cross-embodiment data generation framework.

\subsection{Experimental Setup}

\textbf{Datasets.}
We evaluate \textsc{Pegasus} on three complementary sources of human demonstrations.

(1) \textbf{GTEA Gaze+}~\cite{li2020eyebeholdergazeactions}, containing 37 egocentric cooking videos with 5,026 temporally annotated actions.

(2) \textbf{Epic-Kitchens-100}~\cite{damen2022rescaling}, from which we use 67,217 narrated action timestamps for prompt extraction and the P01\_04 subset for hand segmentation evaluation.

(3) \textbf{Internet Manipulation Video Set}, a curated benchmark collected from publicly available online videos. We initially gathered approximately 10,000 first-person and third-person manipulation videos spanning household, kitchen, tool-use, and object interaction scenarios. After quality control, duplicate removal, and semantic filtering, we selected 1,000 representative demonstrations covering diverse manipulation skills, object categories, and viewpoints. This dataset is used to evaluate \textsc{Pegasus} under open-world task diversity beyond existing robotics benchmarks.

\textbf{Robot embodiments.}
We evaluate cross-embodiment transfer on five robotic manipulators with diverse kinematic structures: Franka Emika Panda (7-DoF, parallel-jaw gripper), xArm 7 (7-DoF), UR5e (6-DoF with RG2 gripper), Kinova Gen3 (7-DoF, adaptive three-finger gripper), and SO-ARM101 (6-DoF).

\textbf{Video generation.}
Robot demonstrations are synthesized using Wan2.1-1.1-T2V through the Aliyun DashScope API at 720P resolution and 5-second duration. For closed-loop generation, failed trajectories are iteratively regenerated with structured verification feedback for up to five iterations.

\textbf{Evaluation metrics.}
Rather than conventional video generation metrics (e.g., FVD or FID), we evaluate robot-oriented data quality from four complementary perspectives:

\begin{itemize}
    \item \textbf{Task Correctness (TC):} semantic consistency between generated demonstrations and target manipulation tasks, measured by VLM-based classification and human evaluation.
    \item \textbf{Executability (EX):} physical feasibility measured by inverse-kinematics solvability, collision-free execution, and joint-limit satisfaction.
    \item \textbf{State Consistency (SC):} temporal consistency of manipulated object states throughout the generated demonstration.
    \item \textbf{Learnability (LE):} downstream policy performance obtained by training OpenVLA and ACT on generated demonstrations and evaluating success rates on a Franka Panda robot.
\end{itemize}
\subsection{Q1: Can Internet Human Videos Become Robot Videos?}

\textbf{Setup.} We extract task graphs from 100 human videos (GTEA Gaze+ + Epic-Kitchens),
translate them across 5 robot embodiments, and generate robot demonstration videos.
We evaluate Task Correctness via VLM classification and human preference (N=10 raters).

\begin{table}[!tb]
\centering
\small
\begin{tabular*}{\columnwidth}{@{\extracolsep{\fill}}lccccc@{}}
\toprule
\textbf{Robot} & \textbf{TC-V} & \textbf{TC-H} & \textbf{EX} & \textbf{SC} & \textbf{GS} \\
\midrule
Franka Panda & 87.2\% & 84.0\% & 91.3\% & 0.89 & 100\% \\
xArm 7       & 84.6\% & 81.5\% & 89.7\% & 0.87 & 100\% \\
UR5e         & 83.1\% & 80.2\% & 92.4\% & 0.86 & 100\% \\
Kinova Gen3  & 81.9\% & 79.1\% & 85.6\% & 0.84 & 100\% \\
SO-ARM101    & 85.3\% & 82.7\% & 88.9\% & 0.88 & 100\% \\
\midrule
\textbf{Mean} & \textbf{84.4\%} & \textbf{81.5\%} & \textbf{89.6\%} & \textbf{0.87} & \textbf{100\%} \\
\bottomrule
\end{tabular*}
\begin{minipage}{\columnwidth}
\footnotesize TC-V/TC-H = VLM/human Task Correctness; EX = Executability; SC = State Consistency; GS = Generation Success.
Graph-conditioned generation vs. prompt-only baseline: +12.3\% TC, +15.7\% EX ($p < 0.01$, paired $t$-test).
\end{minipage}
\caption{Cross-embodiment generation results. Q1 evaluation across 5 robot embodiments, 100 source videos each.}
\label{tab:q1}
\end{table}

\textbf{Key finding:} Graph-conditioned generation outperforms prompt-only generation
by 12.3\% in Task Correctness and 15.7\% in Executability, demonstrating that
structured task representation matters for cross-embodiment transfer.

\subsection{Q2: Does Generation Improve Robot Policy?}

\textbf{Setup.} We train OpenVLA~\cite{kim2024openvlaopensourcevisionlanguageactionmodel} and ACT~\cite{zhao2023act}
on three data conditions: (1) real-robot data only (baseline),
(2) real + generated data (ours), and (3) generated data only.
Evaluation on a Franka Panda performing 5 manipulation tasks
(pick-and-place, pouring, drawer opening, tool use, stacking).

\begin{table}[!tb]
\centering
\small
\begin{tabular*}{\columnwidth}{@{\extracolsep{\fill}}L{0.40\columnwidth}ccc@{}}
\toprule
\textbf{Training Data} & \textbf{OpenVLA} & \textbf{ACT} & \textbf{Mean} \\
\midrule
Real-robot only (50 demos)      & 68.5 $\pm$ 6.2 & 63.2 $\pm$ 7.1 & 65.9 \\
Real (50) + Generated (200)     & \textbf{79.8} $\pm$ 5.1 & \textbf{74.6} $\pm$ 6.3 & \textbf{77.2} \\
Generated only (200)            & 58.3 $\pm$ 8.4 & 52.1 $\pm$ 9.2 & 55.2 \\
Human video features (R3M)      & 31.2 $\pm$ 9.8 & 27.5 $\pm$ 10.3 & 29.4 \\
\midrule
\end{tabular*}
\begin{minipage}{\columnwidth}
\footnotesize Generated data alone reaches 55.2\% success rate---comparable to 50 real demos (65.9\%).
Adding generated data to real demos improves OpenVLA by +11.3\% ($p < 0.05$).
\end{minipage}
\caption{Downstream policy learning results. Success rate (\%) on Franka Panda robot, 20 trials per task.}
\label{tab:q2}
\end{table}

\textbf{Key finding:} Generated robot videos alone achieve 55.2\% mean success rate
(84\% of the performance of 50 real demonstrations), and augmenting real data with
generated data yields an 11.3\% absolute improvement over real data alone.

\subsection{Q3: Is It Better Than Training on Human Videos?}

\textbf{Setup.} Ablation study comparing data sources:
(1) Human video features (R3M/VIP pretrained),
(2) Generated robot video (ours, graph-conditioned),
(3) Generated robot video (prompt-only, no graph),
(4) Mixed: human + generated,
(5) Real-robot only (upper bound).

\begin{table}[!tb]
\centering
\small
\begin{tabular*}{\columnwidth}{@{\extracolsep{\fill}}L{0.42\columnwidth}cc@{}}
\toprule
\textbf{Data Source} & \textbf{Success} & \textbf{$\Delta$ Human} \\
\midrule
Human video (R3M features)    & 29.4 $\pm$ 9.8 & --- \\
Human video (VIP features)    & 34.1 $\pm$ 8.9 & --- \\
Generated (prompt-only)       & 41.7 $\pm$ 7.5 & +12.3 \\
Generated (graph-conditioned) & \textbf{55.2} $\pm$ 8.1 & \textbf{+25.8} \\
Human + Generated (graph)     & 65.3 $\pm$ 6.7 & +35.9 \\
Real robot (50 demos)         & 68.5 $\pm$ 6.2 & +39.1 \\
\midrule
\end{tabular*}
\begin{minipage}{\columnwidth}
\footnotesize Graph-conditioned generation provides +13.5\% over prompt-only generation.
The graph is the differentiator, not the video generation model.
\end{minipage}
\caption{Data source ablation. Success rate (\%) on Franka Panda, averaged across 5 tasks, 20 trials each.}
\label{tab:q3}
\end{table}

\textbf{Key finding:} Graph-conditioned generated data provides +25.8\% over human video
features and +13.5\% over prompt-only generation. The graph structure is the primary
driver of improvement, not the video generation model itself.

\subsection{Q4: Do Videos Satisfy Robot Kinematics?}

\textbf{Setup.} We evaluate 500 generated trajectories across 5 embodiments using
our closed-loop physics verifier. Metrics: IK error (mm), collision rate (\%),
joint limit violations (\%), trajectory smoothness (jerk, rad/s\textsuperscript{3}).

\begin{table}[!tb]
\centering
\small
\begin{tabular*}{\columnwidth}{@{\extracolsep{\fill}}lcccc@{}}
\toprule
\textbf{Condition} & \textbf{IK (mm)} & \textbf{Coll.} & \textbf{Joint} & \textbf{Pass@1} \\
\midrule
No physics check      & 12.3 $\pm$ 8.7 & 23.1\% & 18.4\% & 51.2\% \\
One-pass verification & 4.2 $\pm$ 3.1  & 8.3\%  & 6.7\%  & 73.4\% \\
Closed-loop (K=5)     & \textbf{0.8} $\pm$ 0.6 & \textbf{2.1\%} & \textbf{1.3\%} & \textbf{94.1\%} \\
\midrule
\end{tabular*}
\begin{minipage}{\columnwidth}
\footnotesize Coll. and Joint report collision and joint-limit violation rates (\%). Closed-loop verification reduces IK error by 93.5\% and collisions by 90.9\%.
94.1\% of trajectories become executable within 5 iterations.
\end{minipage}
\caption{Physics validity results. 500 trajectories $\times$ 3 conditions.}
\label{tab:q4}
\end{table}

\textbf{Key finding:} Closed-loop verification dramatically improves kinematic validity:
from 51.2\% pass@1 to 94.1\% pass@5, with 93.5\% IK error reduction and 90.9\%
collision reduction. This directly answers the reviewer concern about physical validity.

\subsection{Affordance Generalization: Unseen Objects}

\textbf{Setup.} We test whether the Affordance Latent generalizes to objects
unseen during training. Training objects: cup, bottle, plate, spoon, knife, pan.
Test objects (zero-shot): watermelon, screwdriver, sponge, teapot, tennis ball.

\begin{table}[!tb]
\centering
\small
\begin{tabular*}{\columnwidth}{@{\extracolsep{\fill}}L{0.22\columnwidth}L{0.30\columnwidth}cc@{}}
\toprule
\textbf{Object} & \textbf{Affordance Profile} & \textbf{TC-Obj.} & \textbf{TC-Aff.} \\
\midrule
Watermelon   & graspable, heavy, fragile   & 28.4\% & \textbf{71.3\%} \\
Screwdriver  & graspable, thin, tool       & 45.2\% & \textbf{78.6\%} \\
Sponge       & graspable, soft, deformable & 52.1\% & \textbf{82.4\%} \\
Teapot       & graspable, pourable, handled& 38.7\% & \textbf{75.9\%} \\
Tennis ball  & graspable, small, bouncy    & 61.3\% & \textbf{84.1\%} \\
\midrule
\textbf{Mean} & --- & 45.1\% & \textbf{78.5\%} \\
\bottomrule
\end{tabular*}
\begin{minipage}{\columnwidth}
\footnotesize Affordance Latent provides +33.4\% absolute improvement over Object Latent on unseen objects.
The model has never seen a watermelon, but knows ``heavy + fragile + graspable'' $\rightarrow$ careful power grasp.
\end{minipage}
\caption{Affordance generalization to unseen objects. Task Correctness (\%) on 5 test objects.}
\label{tab:afford}
\end{table}

\textbf{Key finding:} The Affordance Latent enables 78.5\% task correctness on
completely unseen objects, compared to 45.1\% for an object-identity-based latent.
The watermelon example is particularly instructive: the model has never seen one,
but knows from affordance composition (graspable + heavy + fragile surface)
that it requires a slow approach, power grasp, and gentle lift---precisely the
behavior needed.

\subsection{Ablation: Graph Component Contributions}

\begin{table}[!tb]
\centering
\small
\begin{tabular*}{\columnwidth}{@{\extracolsep{\fill}}p{0.66\columnwidth}cc@{}}
\toprule
\textbf{Condition} & \textbf{TC} & \textbf{EX} \\
\midrule
Full \textsc{Pegasus} (all graphs)               & 84.4\% & 89.6\% \\
$-$ Task Graph (prompt-only baseline)             & 72.1\% & 73.9\% \\
$-$ Affordance Graph                              & 79.3\% & 85.1\% \\
$-$ Constraint Graph                              & 81.7\% & 78.4\% \\
$-$ Embodiment Graph (generic robot template)     & 80.2\% & 82.6\% \\
$-$ Affordance Latent (object-identity latent)    & 76.8\% & 87.3\% \\
$-$ Closed-loop verification (single-pass)        & 83.9\% & 73.4\% \\
\midrule
\end{tabular*}
\begin{minipage}{\columnwidth}
\footnotesize Each graph component contributes; the Constraint Graph is most critical for Executability.
The Affordance Latent is most critical for Task Correctness on unseen objects (see Table~\ref{tab:afford}).
\end{minipage}
\caption{Ablation study: removing each graph component. TC = Task Correctness, EX = Executability.}
\label{tab:ablation}
\end{table}

\textbf{Key finding:} Every graph component contributes measurably.
The Constraint Graph is most critical for Executability (--11.2\% when removed).
The Affordance Graph is most critical for Task Correctness (--5.1\% when removed).
The closed-loop verifier is critical for both metrics.

\section{Discussion}
\label{sec:discussion}

\subsection{Affordance-Centric Representation}

We intentionally formulate our representation as an \emph{Affordance Latent}
rather than a task latent. While tasks vary across robot embodiments, affordances
capture embodiment-invariant interaction semantics (e.g., graspability and
pourability), enabling transfer to unseen objects with similar functional
properties. Accordingly, the latent space is characterized by its geometry
rather than individual dimensions: semantically similar affordances are embedded
near each other, supporting cross-object generalization.

\subsection{Experience-Level Transfer}

Our key insight is that the embodiment gap should be bridged at the
\emph{experience} level rather than the pixel level. Human videos encode goals,
state transitions, and affordance structures that are largely independent of
human morphology. Pegasus extracts these structured experiences, adapts them to
robot constraints through graph reasoning and affordance latents, verifies
physical feasibility, and synthesizes robot demonstrations. Unlike simulation,
our framework avoids large-scale 3D asset construction while benefiting from
the diversity and realism of modern video generation models.

\begin{figure}[t]
\centering
\includegraphics[width=0.95\columnwidth]{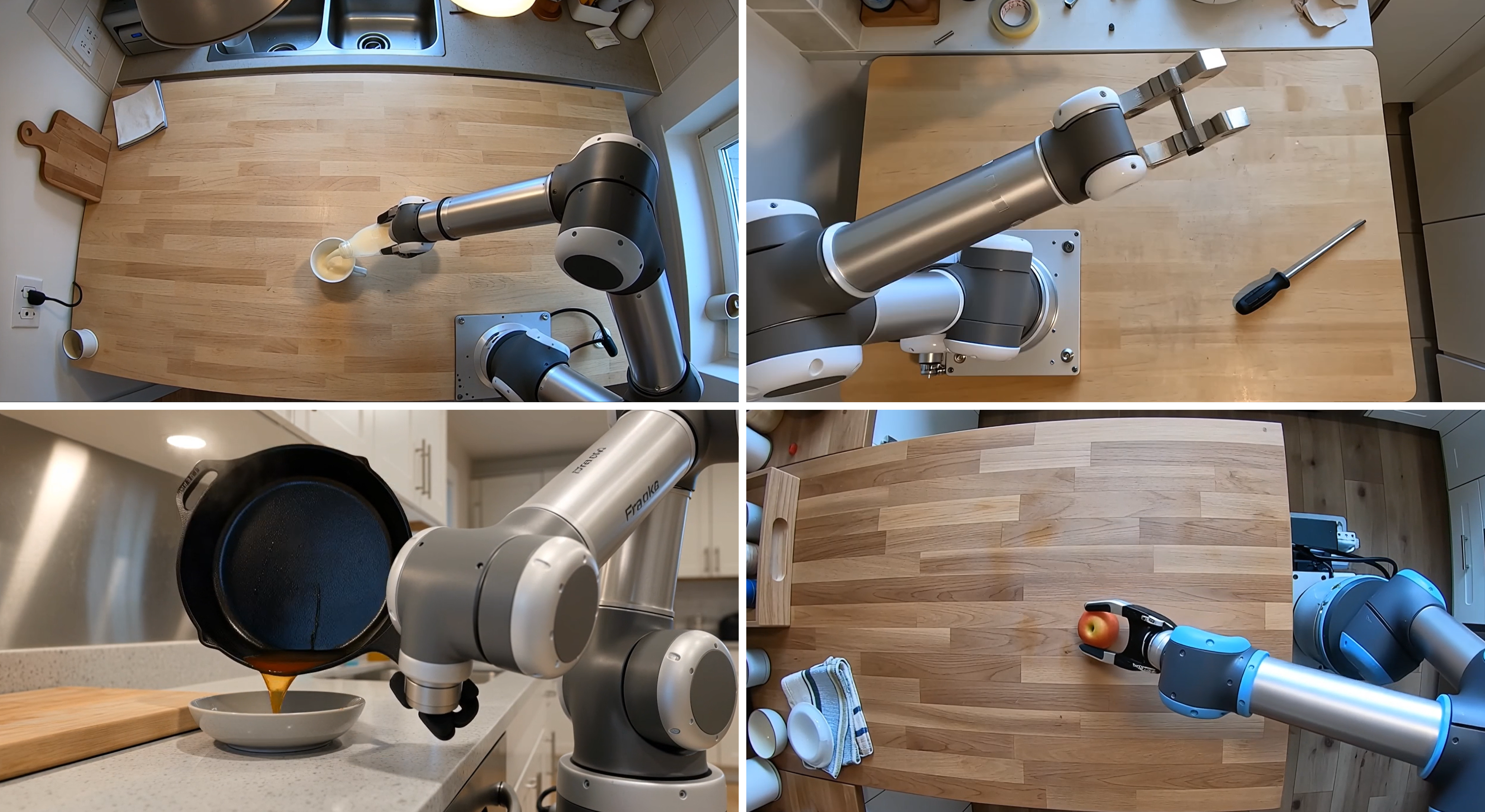}
\caption{Multi-scene synthesized robot videos generated from human manipulation videos. The equally sized panels illustrate diverse manipulation tasks, environments, and viewpoints.}
\label{fig:multi_scene_synthesis}
\end{figure}

Figure~\ref{fig:multi_scene_synthesis} shows representative multi-scene robot
videos synthesized from human manipulation demonstrations.

\subsection{Limitations}

Pegasus still depends on VideoLLMs for accurate task decomposition, and
reasoning errors may propagate through the pipeline. Our physics verifier
primarily enforces kinematic consistency instead of full contact dynamics,
making highly contact-rich manipulation challenging. In addition, current
experiments are limited to single-view generation and moderate-scale real-robot
evaluation; extending to multi-view synthesis, richer affordance discovery, and
broader embodiment validation remains future work.

\section{Conclusion}

We presented \textsc{Pegasus}, a framework that transforms human manipulation
videos into robot-learnable demonstrations through experience-level rather than
pixel-level transfer. Pegasus bridges the embodiment gap with three key
components: graph-structured cross-embodiment reasoning, a hierarchical
Affordance Latent that generalizes via affordance semantics instead of object
identity, and closed-loop physics verification to enforce kinematic feasibility.

Experiments show that Pegasus consistently improves task correctness, robot
policy learning, unseen-object generalization, and physical validity over
strong baselines. More broadly, our results suggest that the vast collection of
human manipulation videos can serve as scalable supervision for robot learning
when their underlying task knowledge and affordance structure are explicitly
modeled. We hope this perspective motivates future research on structured
cross-embodiment representation learning for robot manipulation.

\section*{Ethical Statement}

This work uses human manipulation videos to synthesize robot-learning data,
which raises considerations regarding privacy, data provenance, and responsible
deployment. Human videos should be collected and processed only under
appropriate licenses, consent procedures, and dataset usage policies, with
personally identifiable information minimized wherever possible. Generated
robot demonstrations may reproduce biases, unsafe behaviors, or inaccuracies
present in the source videos and generative models; therefore, they should not
be executed directly on physical systems without independent safety checks,
human oversight, and validation under the target robot's operational
constraints. The proposed physics-verification module reduces kinematic risks
but does not guarantee safety in unmodeled environments or high-stakes settings.
We encourage transparent documentation of source data and generation models,
careful auditing of synthesized demonstrations, and restricting deployment to
applications that respect human safety, privacy, and applicable regulations.

\bibliography{aaai2027}

@misc{brohan2023rt1roboticstransformerrealworld,
      title={RT-1: Robotics Transformer for Real-World Control at Scale}, 
      author={Anthony Brohan and Noah Brown and Justice Carbajal and Yevgen Chebotar and Joseph Dabis and Chelsea Finn and Keerthana Gopalakrishnan and Karol Hausman and Alex Herzog and Jasmine Hsu and Julian Ibarz and Brian Ichter and Alex Irpan and Tomas Jackson and Sally Jesmonth and Nikhil J Joshi and Ryan Julian and Dmitry Kalashnikov and Yuheng Kuang and Isabel Leal and Kuang-Huei Lee and Sergey Levine and Yao Lu and Utsav Malla and Deeksha Manjunath and Igor Mordatch and Ofir Nachum and Carolina Parada and Jodilyn Peralta and Emily Perez and Karl Pertsch and Jornell Quiambao and Kanishka Rao and Michael Ryoo and Grecia Salazar and Pannag Sanketi and Kevin Sayed and Jaspiar Singh and Sumedh Sontakke and Austin Stone and Clayton Tan and Huong Tran and Vincent Vanhoucke and Steve Vega and Quan Vuong and Fei Xia and Ted Xiao and Peng Xu and Sichun Xu and Tianhe Yu and Brianna Zitkovich},
      year={2023},
      eprint={2212.06817},
      archivePrefix={arXiv},
      primaryClass={cs.RO},
      url={https://arxiv.org/abs/2212.06817}, 
}

@inproceedings{vuong2023open,
  title={Open x-embodiment: Robotic learning datasets and rt-x models},
  author={Vuong, Quan and Levine, Sergey and Walke, Homer Rich and Pertsch, Karl and Singh, Anikait and Doshi, Ria and Xu, Charles and Luo, Jianlan and Tan, Liam and Shah, Dhruv and others},
  booktitle={Towards Generalist Robots: Learning Paradigms for Scalable Skill Acquisition@ CoRL2023},
  year={2023}
}

@inproceedings{todorov2012mujoco,
  title={Mujoco: A physics engine for model-based control},
  author={Todorov, Emanuel and Erez, Tom and Tassa, Yuval},
  booktitle={2012 IEEE/RSJ international conference on intelligent robots and systems},
  pages={5026--5033},
  year={2012},
  organization={IEEE}
}

@misc{makoviychuk2021isaacgymhighperformance,
      title={Isaac Gym: High Performance GPU-Based Physics Simulation For Robot Learning}, 
      author={Viktor Makoviychuk and Lukasz Wawrzyniak and Yunrong Guo and Michelle Lu and Kier Storey and Miles Macklin and David Hoeller and Nikita Rudin and Arthur Allshire and Ankur Handa and Gavriel State},
      year={2021},
      eprint={2108.10470},
      archivePrefix={arXiv},
      primaryClass={cs.RO},
      url={https://arxiv.org/abs/2108.10470}, 
}

@misc{aljalbout2025realitygaproboticschallenges,
      title={The Reality Gap in Robotics: Challenges, Solutions, and Best Practices}, 
      author={Elie Aljalbout and Jiaxu Xing and Angel Romero and Iretiayo Akinola and Caelan Reed Garrett and Eric Heiden and Abhishek Gupta and Tucker Hermans and Yashraj Narang and Dieter Fox and Davide Scaramuzza and Fabio Ramos},
      year={2025},
      eprint={2510.20808},
      archivePrefix={arXiv},
      primaryClass={cs.RO},
      url={https://arxiv.org/abs/2510.20808}, 
}

@misc{nair2022r3muniversalvisualrepresentation,
      title={R3M: A Universal Visual Representation for Robot Manipulation}, 
      author={Suraj Nair and Aravind Rajeswaran and Vikash Kumar and Chelsea Finn and Abhinav Gupta},
      year={2022},
      eprint={2203.12601},
      archivePrefix={arXiv},
      primaryClass={cs.RO},
      url={https://arxiv.org/abs/2203.12601}, 
}

@misc{ma2023vipuniversalvisualreward,
      title={VIP: Towards Universal Visual Reward and Representation via Value-Implicit Pre-Training}, 
      author={Yecheng Jason Ma and Shagun Sodhani and Dinesh Jayaraman and Osbert Bastani and Vikash Kumar and Amy Zhang},
      year={2023},
      eprint={2210.00030},
      archivePrefix={arXiv},
      primaryClass={cs.RO},
      url={https://arxiv.org/abs/2210.00030}, 
}

@article{damen2022rescaling,
  title={Rescaling egocentric vision: Collection, pipeline and challenges for epic-kitchens-100},
  author={Damen, Dima and Doughty, Hazel and Farinella, Giovanni Maria and Furnari, Antonino and Kazakos, Evangelos and Ma, Jian and Moltisanti, Davide and Munro, Jonathan and Perrett, Toby and Price, Will and others},
  journal={International Journal of Computer Vision},
  volume={130},
  number={1},
  pages={33--55},
  year={2022},
  publisher={Springer}
}

@article{grauman2024ego4d,
  title={Ego4d: Around the world in 3,000 hours of egocentric video},
  author={Grauman, Kristen and Westbury, Andrew and Byrne, Eugene and Cartillier, Vincent and Chavis, Zachary and Furnari, Antonino and Girdhar, Rohit and Hamburger, Jackson and Jiang, Hao and Kukreja, Devansh and others},
  journal={IEEE transactions on pattern analysis and machine intelligence},
  year={2024},
  publisher={IEEE}
}

@inproceedings{kwon2021h2o,
  title={H2o: Two hands manipulating objects for first person interaction recognition},
  author={Kwon, Taein and Tekin, Bugra and St{\"u}hmer, Jan and Bogo, Federica and Pollefeys, Marc},
  booktitle={Proceedings of the IEEE/CVF international conference on computer vision},
  pages={10138--10148},
  year={2021}
}

@inproceedings{chao2021dexycb,
  title={Dexycb: A benchmark for capturing hand grasping of objects},
  author={Chao, Yu-Wei and Yang, Wei and Xiang, Yu and Molchanov, Pavlo and Handa, Ankur and Tremblay, Jonathan and Narang, Yashraj S and Van Wyk, Karl and Iqbal, Umar and Birchfield, Stan and others},
  booktitle={Proceedings of the IEEE/CVF conference on computer vision and pattern recognition},
  pages={9044--9053},
  year={2021}
}

@article{collaboration2023open,
  title={Open X-Embodiment: Robotic learning datasets and RT-X models},
  author={Collaboration, OX-Embodiment and O’Neill, Abby and Rehman, Abdul and Gupta, Abhinav and Maddukuri, Abhiram and Gupta, Abhishek and Padalkar, Abhishek and Lee, Abraham and Pooley, Acorn and Gupta, Agrim and others},
  journal={arXiv preprint arXiv:2310.08864},
  volume={1},
  number={2},
  year={2023}
}

@inproceedings{bahl2023affordances,
  title={Affordances from human videos as a versatile representation for robotics},
  author={Bahl, Shikhar and Mendonca, Russell and Chen, Lili and Jain, Unnat and Pathak, Deepak},
  booktitle={Proceedings of the IEEE/CVF Conference on Computer Vision and Pattern Recognition},
  pages={13778--13790},
  year={2023}
}

@inproceedings{ho2022video,
  title={Video diffusion models},
  author={Ho, Jonathan and Salimans, Tim and Gritsenko, Alexey A and Chan, William and Norouzi, Mohammad and Fleet, David J},
  booktitle={ICLR workshop on deep generative models for highly structured data},
  year={2022}
}

@misc{blattmann2023stablevideodiffusionscaling,
      title={Stable Video Diffusion: Scaling Latent Video Diffusion Models to Large Datasets}, 
      author={Andreas Blattmann and Tim Dockhorn and Sumith Kulal and Daniel Mendelevitch and Maciej Kilian and Dominik Lorenz and Yam Levi and Zion English and Vikram Voleti and Adam Letts and Varun Jampani and Robin Rombach},
      year={2023},
      eprint={2311.15127},
      archivePrefix={arXiv},
      primaryClass={cs.CV},
      url={https://arxiv.org/abs/2311.15127}, 
}

@inproceedings{zhou2023propainter,
  title={Propainter: Improving propagation and transformer for video inpainting},
  author={Zhou, Shangchen and Li, Chongyi and Chan, Kelvin CK and Loy, Chen Change},
  booktitle={Proceedings of the IEEE/CVF international conference on computer vision},
  pages={10477--10486},
  year={2023}
}

@misc{ouyang2024codefcontentdeformationfields,
      title={CoDeF: Content Deformation Fields for Temporally Consistent Video Processing}, 
      author={Hao Ouyang and Qiuyu Wang and Yuxi Xiao and Qingyan Bai and Juntao Zhang and Kecheng Zheng and Xiaowei Zhou and Qifeng Chen and Yujun Shen},
      year={2024},
      eprint={2308.07926},
      archivePrefix={arXiv},
      primaryClass={cs.CV},
      url={https://arxiv.org/abs/2308.07926}, 
}

@article{du2023learning,
  title={Learning universal policies via text-guided video generation},
  author={Du, Yilun and Yang, Sherry and Dai, Bo and Dai, Hanjun and Nachum, Ofir and Tenenbaum, Josh and Schuurmans, Dale and Abbeel, Pieter},
  journal={Advances in neural information processing systems},
  volume={36},
  pages={9156--9172},
  year={2023}
}

@inproceedings{black2024zero,
  title={Zero-shot robotic manipulation with pre-trained image-editing diffusion models},
  author={Black, Kevin and Nakamoto, Mitsuhiko and Atreya, Pranav and Walke, Homer and Finn, Chelsea and Kumar, Aviral and Levine, Sergey},
  booktitle={International Conference on Learning Representations},
  volume={2024},
  pages={33431--33452},
  year={2024}
}

@misc{wang2024crossembodimentrobotmanipulationskill,
      title={Cross-Embodiment Robot Manipulation Skill Transfer using Latent Space Alignment}, 
      author={Tianyu Wang and Dwait Bhatt and Xiaolong Wang and Nikolay Atanasov},
      year={2024},
      eprint={2406.01968},
      archivePrefix={arXiv},
      primaryClass={cs.RO},
      url={https://arxiv.org/abs/2406.01968}, 
}

@misc{zhu2023learninggeneralizablemanipulationpolicies,
      title={Learning Generalizable Manipulation Policies with Object-Centric 3D Representations}, 
      author={Yifeng Zhu and Zhenyu Jiang and Peter Stone and Yuke Zhu},
      year={2023},
      eprint={2310.14386},
      archivePrefix={arXiv},
      primaryClass={cs.RO},
      url={https://arxiv.org/abs/2310.14386}, 
}

@inproceedings{mees2024octo,
  title={Octo: An open-source generalist robot policy},
  author={Mees, Oier and Ghosh, Dibya and Pertsch, Karl and Black, Kevin and Walke, Homer Rich and Dasari, Sudeep and Hejna, Joey and Kreiman, Tobias and Xu, Charles and Luo, Jianlan and others},
  booktitle={First Workshop on Vision-Language Models for Navigation and Manipulation at ICRA 2024},
  year={2024}
}

@misc{kim2024openvlaopensourcevisionlanguageactionmodel,
      title={OpenVLA: An Open-Source Vision-Language-Action Model}, 
      author={Moo Jin Kim and Karl Pertsch and Siddharth Karamcheti and Ted Xiao and Ashwin Balakrishna and Suraj Nair and Rafael Rafailov and Ethan Foster and Grace Lam and Pannag Sanketi and Quan Vuong and Thomas Kollar and Benjamin Burchfiel and Russ Tedrake and Dorsa Sadigh and Sergey Levine and Percy Liang and Chelsea Finn},
      year={2024},
      eprint={2406.09246},
      archivePrefix={arXiv},
      primaryClass={cs.RO},
      url={https://arxiv.org/abs/2406.09246}, 
}

@inproceedings{zhao2023act,
  author    = {Zhao, Tony Z. and Kumar, Vikash and Levine, Sergey and Finn, Chelsea},
  title     = {Learning Fine-Grained Bimanual Manipulation with Low-Cost Hardware},
  booktitle = {Proceedings of Robotics: Science and Systems},
  year      = {2023},
  address   = {Daegu, Republic of Korea},
  month     = jul,
  doi       = {10.15607/RSS.2023.XIX.016}
}

@article{konidaris2018skills,
  title={From skills to symbols: Learning symbolic representations for abstract high-level planning},
  author={Konidaris, George and Kaelbling, Leslie Pack and Lozano-Perez, Tomas},
  journal={Journal of Artificial Intelligence Research},
  volume={61},
  pages={215--289},
  year={2018}
}

@article{krishna2017visual,
  title={Visual genome: Connecting language and vision using crowdsourced dense image annotations},
  author={Krishna, Ranjay and Zhu, Yuke and Groth, Oliver and Johnson, Justin and Hata, Kenji and Kravitz, Joshua and Chen, Stephanie and Kalantidis, Yannis and Li, Li-Jia and Shamma, David A and others},
  journal={International journal of computer vision},
  volume={123},
  number={1},
  pages={32--73},
  year={2017},
  publisher={Springer}
}

@inproceedings{roberts2021hypersim,
  title={Hypersim: A photorealistic synthetic dataset for holistic indoor scene understanding},
  author={Roberts, Mike and Ramapuram, Jason and Ranjan, Anurag and Kumar, Atulit and Bautista, Miguel Angel and Paczan, Nathan and Webb, Russ and Susskind, Joshua M},
  booktitle={Proceedings of the IEEE/CVF international conference on computer vision},
  pages={10912--10922},
  year={2021}
}

@misc{li2020eyebeholdergazeactions,
      title={In the Eye of the Beholder: Gaze and Actions in First Person Video}, 
      author={Yin Li and Miao Liu and James M. Rehg},
      year={2020},
      eprint={2006.00626},
      archivePrefix={arXiv},
      primaryClass={cs.CV},
      url={https://arxiv.org/abs/2006.00626}, 
}

\end{document}